# Cognitive Analysis of 360 Degree Surround Photos

## An Augmented Image Processing Technique for Photosphere Analysis

Madhawa Vidanapathirana, Lakmal Meegahapola, Indika Perera
Department of Computer Science and Engineering
University of Moratuwa
Sri Lanka
{madhawa.13, lakmalbuddikalucky.13, indika}@cse.mrt.ac.lk

*Abstract*—360 degrees surround photography or photospheres have taken the world by storm as the new media for content creation providing viewers rich, immersive experience compared to conventional photography. With the emergence of Virtual Reality as a mainstream trend, the 360 degree photography is increasingly important in offering a practical approach for the general public to capture virtual reality ready content from their mobile phones without explicit tool support or knowledge. Even though the amount of 360 degree surround content being uploaded to the Internet continues to grow, there is no proper way to index them or to process them for further information. This is because of the difficulty in image processing the photospheres due to the distorted nature of objects embedded. This challenge lies in the way 360 degree panoramic photospheres are saved. This paper presents a unique, and innovative technique named Photosphere to Cognition Engine (P2CE), which allows cognitive analysis on 360 degree surround photos using existing image cognitive analysis algorithms and APIs designed for conventional photos. We have optimized the system using a wide variety of indoor and outdoor samples and extensive evaluation approaches. On average, P2CE provides up-to 100% growth in accuracy on image cognitive analysis of photospheres over direct use of conventional non-photosphere based Image Cognition Systems.

## I. INTRODUCTION

After the invention of photography in 19th century, the desire of photographers have grown into finding ways to show more content or a richer view of cities and landscapes. This paved the way for wide angle lenses, which could capture almost up to 180 degree views in the surrounding. As shown by Fig. 1 people started capturing 180 degree panoramas [1]. This development later led to capturing of 360 degree panoramas as shown by Fig. 2 [2]. These panoramas give a 360 degree horizontal view of what is around you and could be considered as an extension of 180 degree panoramas.

With the advancements in image processing and hardware technology, now we are in an age where the latest mode of photography is 360 degree surround photography or photospheres as shown in Fig. 3. Photospheres let users capture the entire viewable surrounding from a point. This gives a full 360° view horizontally as well as a full 180° view vertically, which is what normal people are used to see in real life, with the freedom of rotational movement of head and body. This could also be named as a seamless view. It allows users to capture the entire surrounding of where the user is. The photos are saved as JPG or PNG with dimensions width to height being in the ratio 2:1. Android smart phones already have the built in capability to capture photospheres [3] and immergence of 360 cameras such as Gear 360 [4] and Panono [5] has made the process of capturing photospheres even easier.

With Facebook [6], [7] introducing 360 degree surround photos to the most used social network in the world, it has become a trend now that many people capture 360 surround photos with their phones and upload to Facebook. These photospheres can even be viewed with Virtual Reality (VR) headsets giving users unique, immersive experience. Hence for the general public who do not have programming or thorough technical knowledge on creating virtual reality content, taking photospheres has become the easiest and the most comprehensive way to create VR content. Present day tech giants such as Facebook and Google with their Virtual Reality platforms: Oculus and Day Dream have understood these phenomena and have facilitated users to view, create and share 360 surround photos in their respective platforms.

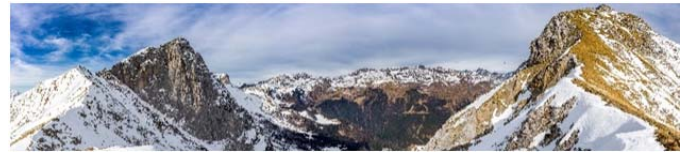

Fig. 1.    180 degree panorama.

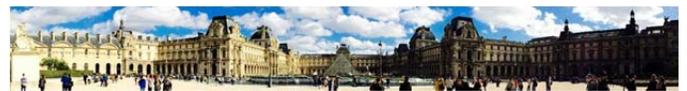

Fig. 2.    360 degree panorama.

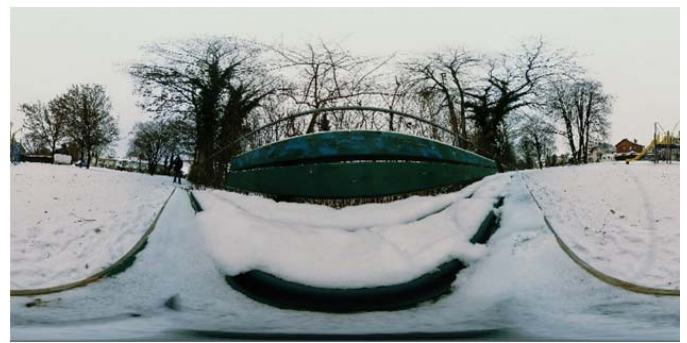

Fig. 3.    Full 360 degree surround photo or photosphere.

On the other hand we have cognition, which is defined as the mental action or process of acquiring knowledge and understanding through thought, experience, and the senses. With the technological advancements in the fields of Artificial Intelligence (AI) and Image Processing, software companies are in a quest to build applications and platforms that are intelligent. Microsoft with their cognition platform Microsoft Cognitive Services [8] and Google with their Vision APIs [9] are trying to give cognition to photos we see every day. A practical example of this can be seen in Facebook such that, when we upload photos, it automatically recognizes people in the photo. This is in fact a Face Detection [10] and Face Recognition [11] Problem, which could be identified as an early stage cognition challenge. Today, image cognition is rather focused on Object Detection [12] and computer generated wholes-tic explanations of images. This combination between cognition and photos can be seen in numerous other places too, making it an interesting area of research to explore.

Because 360 degree surround photos are a new area of research, ample voids exist when compared to conventional photography related research. It is observed that currently available cognition platforms with photospheres do not give accurate results because of the way photospheres are saved and the distortions in the objects and entities present in the photospheres. Even in Facebook, they do not allow people to be recognized in photospheres through their cognition/face-recognition engines. Hence there is a need for a solution addressing this challenge by augmenting currently available cognition platforms with the ability to process photospheres.

This paper presents a solution with an optimized methodology to process photospheres through currently available image processing platforms often used for cognition, optical character recognition [13], content moderation and many other purposes. The solution is known as P2CE or Photosphere to Cognition Engine. Section II outlines the related work that has been carried out in the area of interest. Section III describes the proposed methodology and Section IV explains experiments we carried out. Section V discusses the results of the experiments. Finally, Section VI concludes the paper with possible future research directions.

## II. BACKGROUND AND RELATED WORK

Delving deep into Rectifiers: Surpassing human-level performance on ImageNet classification [14] by Microsoft research discusses the cognitive framework behind Microsoft Cognitive Services, which uses rectified activation units that are essential for state-of-the-art neural networks. Using these techniques this methodology is the first to surpass human level performance on visual recognition challenge on ImageNet 2012 classification data-set. That data-set contains planar images and authors have not provided details on how processing can be done on 360 degree surround photos. Further this paper discusses how cognition is important to build intelligent applications.

Convolutional Feature Masking for Joint Object and Stuff Segmentation [15] by Microsoft research analyses the use of semantic segmentation using convolutional neural networks for accurate feature recognition of planar images. In this research they have exploited shape information by masking convolution features. They have further gone on to improve the way of recognizing objects and "stuff" such as grass, sky, water using the above methodology. In their paper, they emphasize the significance of identifying scenes while giving cognition to the outdoor scenes.

Rendering 3D tessellations with conformal curvilinear perspective [16] discusses how perspective of paintings project the world from the point of our eyes on to a single rectangular screen (similar to normal images) in contrast to spherical images created by projecting world onto a closed sphere surrounding the eye-point, proposing the concept called viewable sphere. This is the exact concept used in Virtual Reality and in gaming industry as well when using skybox. The paper further discusses the Stereographic projection, which is a projection obtained when the eye of a person sitting in the center of a sphere reaches the surface of the sphere itself. Then authors go on to discuss the concept of Cube Maps generated from the viewable photospheres. Geometric based stereoscopic rendering when dealing with virtual environments with wide field of views up to 360 degrees has been discussed in Stereoscopic Rendering of Virtual Environments with Wide Field-of-Views up to 360 [17]. These were sources of inspiration for the research experiment discussed in this paper as well.

Jump: Virtual Reality Video [18] by Google researchers identify why capturing content for virtual reality has become so important. It considers the fact that VR headsets becoming widely available all over the world and how this area of study has led to many new research problems. Further they have discussed about Omnidirectional Stereo (ODS) view, which is a basic principle in 360 surround photography. This view allows users to look around a 360 view but without moving. Here they mention the fact that most suitable representation for 360 photospheres and videos is the use of planar images in width to height ratio 2:1 as shown in Fig. 3 as well.

Another research done in Google, "Large-scale Privacy Protection in Google Street View" [19] discusses how they have approached the identification of license plates and faces in 360 photospheres used in Google Street View to blur them automatically. Here they wanted to recognize two things in images. They do not want any other large objects available in the photosphere to be recognized. Hence the approach they have taken is to create the datasets used for recognition in a manner so that they account for distortions and effects due to the curvy nature of faces and license plates (due to the way photospheres are saved). Since their scope is limited to identifying (detecting, not recognizing) license plates and faces, generating new data-sets has been viable. However, taking into account distortions and mishaps in all the objects in the world and creating a dataset specifically to recognize features in 360 photospheres is a very difficult task to do. Additionally, the use of cognition models trained for conventional photos for cognition on 360 surround photos terribly fail as shown by our test results in Fig. 9 and 10.

Putting Objects in Perspective [20] describes how humans percept differently when an image is on 2D image plane where it was projected by the camera on a plane (normal images we see) and within the real 3D world (which is what 360 photospheres try to capture) where it is all like a projection on a sphere around us. Further this research paper discusses how computer vision deals with Perspective Projection and regarding semantic interpretation of an image as well. Nevertheless, the paper does not go in to describing mechanisms to do image processing to 360 degree projections as the main research area is different.

Recognizing scene view point using panoramic place representation in [21] presents a way to classify the type of place shown in the planer photo and also to recognize observer's viewpoint within that category of place. In that research they have used 360 photospheres as the training datasets because 360 degree photospheres give coverage of all possible viewports within a particular place. Researchers have used a mechanism where they get a conventional photograph as test image and compare it with the 360 photospheres (training data) using 12 different view ports (of each training image) to get a match. The approach they have taken is not able to cognitively analyze photospheres. After all, what they present is another cognitive analysis platform similar to Microsoft's Cognitive Services or IBM Watson Visual Recognition APIs [22] where they use 360 photospheres as the training dataset.

## III. METHODOLOGY

### A. Overview

The methodology we propose as P2CE can be used to convert a photosphere to a set of square dimensioned images that are optimized to be compatible with typical cognitive image analysis systems such as Google Vision API and Microsoft Cognitive Services API. The approach taken by us is inspired by the concept of Cube-Maps, which is commonly used in 3D computer graphics to generate reflection and sky-boxes [23].

A cube map is a collection of six squared images that represent the six sides of an axis aligned cube. By placing a camera at a fixed location and rotating it to face six orthogonal directions, one can obtain six photographs as shown in Fig. 4. These six photos represent a Cube-Map with each photo corresponding to a side of the cube. The camera used should have 90° Field of View in both Horizontal and Vertical Axis. The images represented by a Cube-Map can be mapped to a sphere (as in Fig. 5) and if the users viewpoint is placed at the center inside the sphere, the user would observe the photosphere represented by the Cube-Map.

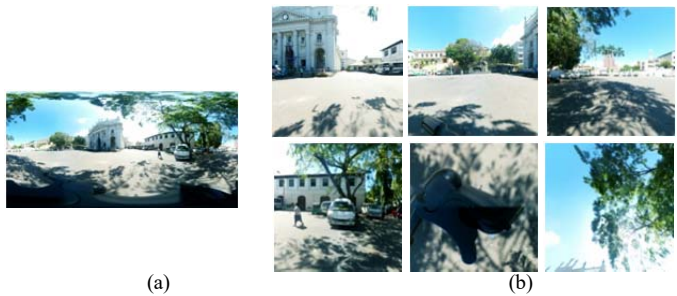

(a)       (b)

Fig. 4. (a) Represents a raw 2:1 photosphere, (b) Represents the 6 images which can be obtained after converting the photosphere to a cube map.

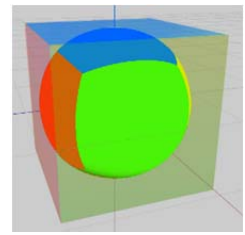

Fig. 5. How a cube-map and a sphere aligns together.

As explained previously, a Cube-Map can represent a Photosphere; additionally, by processing the photos of the sides of a Cube-Map through a typical Image Cognitive Analysis API, it is possible to obtain reasonably accurate results on content of the Photosphere represented by the Cube-Map. In this process, the images corresponding to top side and bottom side of Cube Map can be ignored since in most cases, they do not include additional useful information. The sky and the ground are already visible in the lateral side images of Cube-Map.

Directly feeding 2:1 photosphere images into typical Image Cognitive Analysis APIs yield inaccurate results due to curvy distortion caused by high field of view. Even though the most intuitive approach is to covert the 2:1 image into a Cube-Map as explained before, it is not an optimal solution (as shown by our test results later) because:

- Cube-Map faces represent 90° field of view (FOV), which still demonstrates a higher degree of distortion than a conventional photo.

- Objects that get trimmed by edges of Cube-Map face images are not properly identifiable by typical image cognitive analysis systems.

In this paper, we answer the above two problems and identify the optimal solution to convert a 2:1 photosphere to a set of photos that are highly compatible with typical Cognitive Analysis systems. We generalize the idea of Cube-Map into n-gonal Regular Prism Maps (hereby referred to as n-gonal Prism Maps or Prism Maps) and propose a technique to convert a 2:1 photosphere image into lateral sides of a vertical axis aligned n-gonal regular prism. Through our testing, we obtained the optimal *n* value and *field of view (FOV)* angle for best precision and recall from typical image cognitive analysis systems.

### B. Vertical Axis Aligned Prism Maps

A prism, as shown in Fig. 6(b), is the extrusion of a polygon that represents its base (B). An extrusion extends a polygon into the 3rd dimension by converting its side into equal length (h) rectangular planes that are normal to polygon surface. Effectively, it connects two similar polygons in two different planes.

The "n" of n-gonal prism represents the number of sides of polygon used as the base of prism. Sometimes we use common names to replace "n-gonal" such as Pentagonal and Hexagonal to form Pentagonal Prism and Hexagonal Prism.

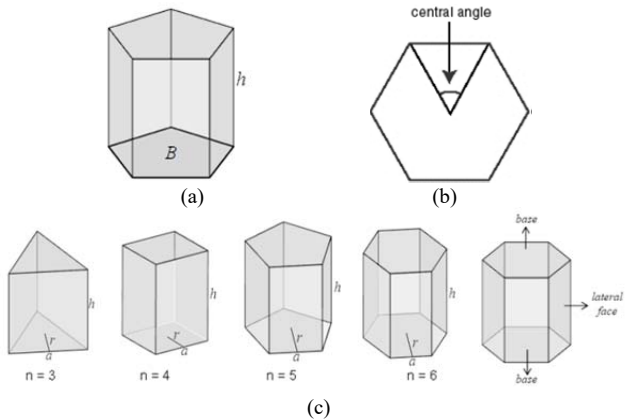

Fig. 6. (a) Hexagonal Prism. The base and height of the prism is indicated. (b) Hexagon. Shows how the central angle is measured (c) Shows how n-gonal prism changes when n is changed.

The rectangular sides that interconnect the two bases of a prism are known as the lateral faces or lateral sides. A regular prism is a prism having a regular polygon as its base. A regular polygon is a polygon with all sides having equal length. As indicated in Fig. 6(b), the angle measured between two adjacent vertices of a polygon at its center is known as the central angle. The central angle of a regular polygon is constant for all consecutive vertex pairs. The central angle (C) for a regular n-sided polygon can be calculated by dividing 360 by n.

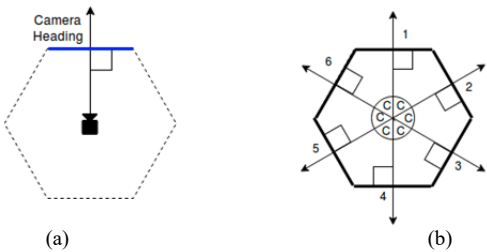

Fig. 7. (a) Camera heading at outward normal direction, (b) Headings of photos taken for lateral sides of hexagonal prism.

The images corresponding to the n lateral sides of a Vertical Axis aligned Prism Map can be obtained by taking photos from cameras placed at center of the Prism, heading at the outward normal directions (Fig. 7(a)) of corresponding lateral sides. In total, there would be n photos taken with angle between the headings of two consecutive photos being equal to central angle of the base of prism as shown in Fig. 7(b).

### C. Converting 2:1 Photosphere to Lateral Sides of Prism Map

Based on our inspiration from Cube Maps, we have continued to use equal vertical and horizontal field of view angles to capture images for lateral sides of Regular Prism Map. Thus, the images obtained are square images. Additionally, we have ignored the top and bottom faces as they do not contain additional information. The square dimensioned images of the lateral sides of a Prism Map can be obtained from a 2:1 photosphere using Perspective Projection [24] of a 3D Scene as described in following steps:

*1)* Define an empty scene in 3D space with positive Z direction being analogous to vertical up direction.

*2)* Create a hollow sphere (S) with radius R, center aligned with origin of scene.

*3)* Apply the Photosphere as the inner surface material of hollow sphere.

*4)* Place a perspective projection [24] camera (C) with same vertical and horizontal Field of View (FOV). The camera should be looking at positive X direction with Up-side in positive Z direction.

*5)* Render the scene captured by C on a square dimensioned image using a shading algorithm which does not consider lighting. In other words, the output should be similar to a scene lightened by a pure white Ambient Light.

*6)* Rotate the camera about Z axis by an angle 360/n and Repeat from Step 5, until the camera completes a full rotation.

*7)* Combine each squared image obtained in Step 5 and generate an n-gonal Prism Map.

The pseudo-code for the n-gonal Prism Map conversion algorithm stated above is shown in Algorithm 1.

---

**Algorithm 1** Photosphere to n-map conversion

**Require:** 360 photosphere
1.   **Function** *photophereToNMap*(photosphere,n,FOV)
2.       NMAP = new ngonPrismMap(n)
3.       SCENE = *CreateScene*(photosphere,FOV)
4.       set C = new *PerspectiveProjectionCamera*(FOV)
5.       S.add(C,Position=<0,0,0>,Heading=<1,0,0>, UpDirection=<0,0,1>)
6.       For i=1 to n
7.           set image = C.render(SCENE, Lighting =  DISABLED)
8.           set NMAP.LateralSides[i] = image
9.           C.rotate(360/n degrees, axis = Z axis)
10.      Repeat
11.      **return** NMAP
12.  **Function** *CreateScene*(photosphere,fov)
13.      set R = 1000 //Any sufficiently large value
14.      set SCENE = new empty 3D scene with positive Z as vertical up axis
15.      set S = hollow sphere with radius R
16.      SCENE.add(S, Position=<0,0,0>)
17.      set S.InnerSurfaceTexture = photosphere
18.      **return** SCENE

---

Given below is the terminology of Algorithm 1:

• Function photosphereToNMap converts a photosphere to a Prism Map.

• CreateScene – Creates the 3D scene setup used for rendering of Prism Map Sides.

• SCENE.add – Adds a 3D primitive to the scene.

• S.InnerSurfaceTexture – The image/texture assigned to interior surface of sphere S.

• PerspectiveProjectionCamera – Captures the scene and render to output using perspective projection.

• Heading – Direction in which camera is looking at.

• UpDirection – Direction in which up-side of output photos are heading. Determines roll of camera.

- Render – Renders the scene to image, with lighting disabled. (Similar output to placing a pure white ambient light.)

- C.rotate – Rotate the camera about origin of Scene.

The Mapping of 2:1 Photosphere to the Hollow Sphere should be done as a Cylindrical Projection, similar to how a planner world map is mapped to the globe as shown in Fig. 8. Table 1 demonstrates the conversion of 2:1 Photosphere in Fig. 9 [25] to Lateral Sides of a Hexagonal Prism Map.

### D. Processing n-gonal Prism Maps for Cognition

The images of lateral sides of Prism Maps can be processed using typical Image Cognitive Analysis APIs to obtain reasonable results. The *field of view* used to create the n-gonal Prism Maps is lesser than that of typical Cube Maps and hence the Image Distortions are less.

Additionally, by choosing a higher *field of view* than *Central Angle* (FOV > C), errors due to trimming at edges can be avoided. Thus, the recall and precision of Cognitive Analysis is improved when compared to processing Cube Maps or directly feeding 2:1 Photosphere to Cognition Engine.

F1 Score is a statistical measure of accuracy in classifications. Through experiments, we have calculated the optimal value for *Field of View (FOV)*, *Central Angle (C)* and *number of sides (n)* in order to obtain maximum F1 score.

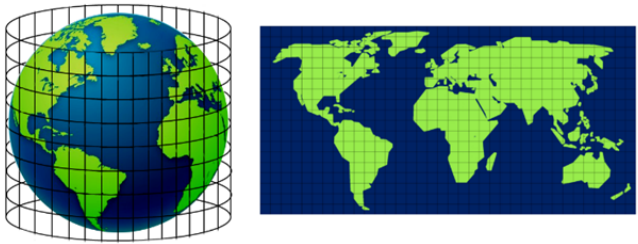

Fig. 8. Cylindrical projection of a sphere.

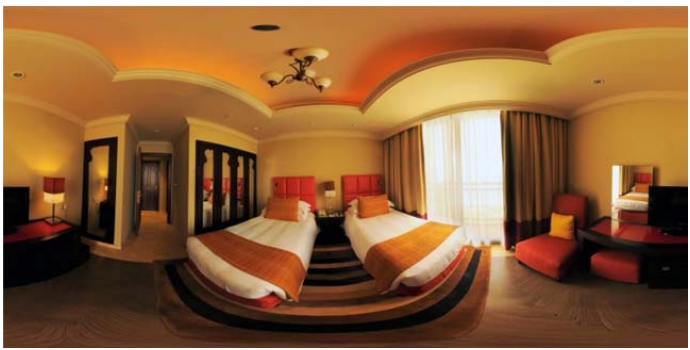

Fig. 9. 2:1 Photosphere which is used for the test.

TABLE I.     HEXAGONAL PRISM MAP IMAGES GENERATED FROM OUR METHODOLOGY USING THE PHOTOSPHERE IN FIG. 9.

| # | Scene (Top Down View) | Hexagonal Prism Map Image |
|---|---|---|
| 1 | | |
| 2 | | |
| 3 | | |
| 4 | | |
| 5 | | |

| # | Scene (Top Down View) | Hexagonal Prism Map Image |
|---|---|---|
| 6 | 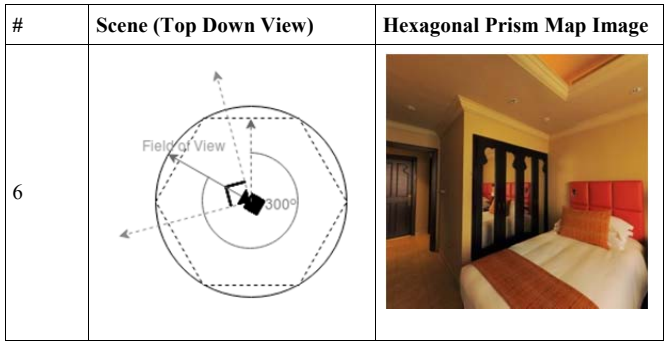 | |

TABLE II.    CONFIGURATIONS OF PRISM MAP USED IN EXPERIMENT

| n : Central Angle | Field of View Angles |
|---|---|
| 3 : 120 | 120 |
| 4 : 90 | 90, 120 |
| 6 : 60 | 60, 90, 120 |
| 8: 45 | 45, 52, 60, 90, 120 |

For each 2:1 Photosphere in test set, 11 Prism Maps were obtained following each configuration in the table above (Table 2). Afterwards, n lateral side images of each n-gonal Prism Map were processed using Cognitive Analysis APIs in order to obtain (Label, Confidence) tuples that describe them.

Afterwards, based on a predefined threshold Confidence value $T_x$, the labels with Confidence > $T_x$ are added to a separate list of labels known as *positives*. For example, if our threshold confidence is $T_{75}$ and we obtain the label "Outdoor" with confidence 0.65, then we do not add "Outdoor" to *positives* list. The confidence given by the system is not sufficient for us to be sure that the image indeed contains outdoor scenery.

To further clarify, it could be stated as follows. For a given $T_x$ and for a considered configuration of our methodology, each Prism Map (of a Photosphere) would have its own *positives* list containing labels with *confidence* > $T_x$ obtained via Cognitive Analysis of images of its sides.

To calculate the precision and recall of a label, we need to obtain a list of negative labels as well. However, when defining negative labels, we have to ensure that the corresponding label is known by the Cognitive Analysis System used. For example, we could define a negative label "T Shirt" but if the cognitive analysis system used does not include the word "T Shirt" in its vocabulary, then our test would not properly measure the accuracy of our methodology. It would rather measure the inability of cognitive analysis system used. Therefore, we have considered the union of all *positives* lists for a given confidence of the considered photosphere as the list of words in *vocabulary* (1) used to process the photosphere. This list would contain labels generated for all 11 configurations of methodology used by us. The list of *negatives* (2) for a given photosphere can then be formed by removing elements in *positives* from *vocabulary* list.

The idea behind separating labels to *positives* and *negatives* lists is to let us calculate precision and recall figures for each photosphere at a given threshold $T_x$. By separating labels into *positives* and *negatives*, we can define *True Positives* (3), *False Positives* (4) and *False Negatives* (5) which then can be used to calculate *Precision* (6), *Recall* (7) and *F1* (8) values.

$$\text{Vocabulary} = \cup_{i=1}^{11} \text{positives}_i \qquad (1)$$

$$\text{Negatives}_i = \{x : (x \in \text{Vocabulary}) \text{and} (x \notin \text{Positives}_i)\} \qquad (2)$$

## IV. EXPERIMENTAL SETUP

The experiments were conducted on a computer which was running Ubuntu 16.04 64-bit operating system. The system used an Intel i7 5700 HQ 2.7 GHz Quad Core Hyper-threaded Mobile Processor (Turbo Clock 3.5 GHz). The system consisted of 8 GB Memory with 1 TB storage space. No dedicated Graphics Cards were used.

The bottleneck for the entire experiment can be identified as the time consumed to feed the Prism Map lateral side images to Cognitive Analysis Servers. At 1024x1024 resolution and 32 bit color depth, a PNG face consumes about 1 MB bandwidth. At same configuration, a JPEG face consumes about 200KB. Both configurations provide similar results from Cognitive Analysis Systems.

We have used blender to render the 3D scene with Python Scripts to support the implementation. Microsoft Cognitive Analysis Vision API and Google Vision API were used to test the accuracy of our methodology. Both aforementioned systems allow generation of (Label, Confidence) tuples that describes an image. A label is usually a 1 word text such as "Outdoor", "Water" or "Crowd" that would explain the image. The confidence is a percentage value that indicates how accurate the cognitive engine is in determining the label.

## V. EVALUATION, RESULTS & DISCUSSION

We have prepared 32 photosphere images containing indoor and outdoor locations as our test set. The photospheres were processed using our methodology via Microsoft Cognitive Analysis Vision API and Google Vision API in order to obtain (Label, Confidence) tuples describing the photospheres. Different configurations of our methodology were used with variations in *FOV* and *n* and the variation of Recall and Precision were studied.

Through our experiments, it was understood that by using our methodology, the F1 scores would approximately double when compared to directly feeding the 2:1 Photosphere Image to typical Cognitive Analysis APIs.

### A. Experimental Procedure

We have considered following 11 configurations of Prism Maps in our experiment:

$$TruePositives_i =$$

$$\{x: (x \in Positives_i) and (_{x} \overset{\text{based on human inspection}}{\text{correctly describes the photosphere}})\} \quad (3)$$

$$FalsePositives_i =$$

$$\{x: (xPositives_i) and (_{x} \overset{\text{based on human inspection}}{\text{doesn't correctly describe the photosphere}})\} \quad (4)$$

$$FalseNegatives_i =$$

$$\{x: (x \in Negatives_i) and (_{correctly describes the photosphere}^{\text{based on human inspection x}})\} \quad (5)$$

Define Cardinality of X as count (X)

$$Precision_i = \frac{count(TruePositives_i)}{count(TruePositives_i) + count(FalsePositives_i)} \quad (6)$$

$$Recall_i = \frac{count(TruePositives_i)}{count(TruePositives_i) + count(FalseNegatives_i)} \quad (7)$$

$$F1_i = \frac{2 \times Precision_i \times Recall_i}{Precision_i + Recall_i} \quad (8)$$

As explained previously, using the *positives* and *negatives* label lists, we can calculate *precision*, *recall* and *F1* values for generated labels of each photosphere under different configurations of our methodology for different thresholds $T_x$. For our experiment, we have chosen Thresholds as $T_{75}$, $T_{50}$ and $T_{25}$. By considering all 32 samples in Test Set, we obtained average precision, average recall and average F1 for each threshold. The corresponding standard deviation values are also obtained.

The entire experiment procedure mentioned above is explained using pseudo-code in Algorithm 2. Given below is the terminology of Algorithm 2:

- $T_x$ is the threshold fraction used to identify positives.

- TestSet – List of 2:1 Photosphere images in Test Set.

- ConfigurationsArray - List of Configurations to use. Configuration.FOV = Field of View, Configuration.n = number of lateral sides of Prism Map.

- photosphereToNMap – Our methodology which is defined in Algorithm 1.

- obtainLabels – Process an image through Cognitive Image Analysis API used and obtain (label, confident) pairs.

- NMAP – n-gonal Prism Map as defined in Algorithm 1.

- Vocabulary – List of words assumed to be associated with processed sample photosphere by the Cognitive Analysis API used.

- Truth – Subset of words from Vocabulary, that a human would identify as proper descriptions of considered sample photosphere.

| Algorithm 2 |
|---|
| **Require:** 360 photosphere |
| 1.   ***Function Experiment***(Tx,TestSet,ConfigurationsArray) |
| 2.      Foreach sample in TestSet |
| 3.        set positive_lists = {} |
| 4.        Foreach configuration in ConfigurationsArray |
| 5.          NMAP = photosphereToNMap(sample, configuration.n, configuration.FOV) |
| 6.          set positives = {} |
| 7.          Foreach image in NMAP.LateralSides |
| 8.            set labels = obtainLabels(image) |
| 9.            Foreach label in labels |
| 10.              if label.confidence > Tx |
| 11.                positives.add(label) |
| 12.          positive_lists.put(configuration, positives) |
| 13.        set vocabulary={x:x ∈ P for some P ∈ positives_list} |
| 14.        set truth = {x: x ∈ vocabulary and x describes sample according to human inspection} |
| 15.        Foreach config in ConfigurationsArray |
| 16.          set positives = positive_lists[config] |
| 17.          set false_negatives = truth – positives |
| 18.          set true_positives= truth ∩ positives |
| 19.          set false_positives= positives - truth |
| 20.          Calculate precision, recall and F1 using true_positives, true_negatives, and false_negatives |
| 21.      **return** average and standard deviation of precision, recall and F1. |

### B. Experiment Results & Discussion

Directly processing the sample 2:1 photospheres without P2CE has a F1 of only about 30% - 40% in both Google Vision API and Microsoft Cognitive Services Vision API (as shown in Fig. 10 and 11). This is due to the significantly less recall figures of 16% - 25% shown by both the APIs for directly processing our test set. However, the best configurations of P2CE increased the F1 figures up-to 70% - 80% range. This is a two-fold increment in F1 figures. The recall figures have also increased up-to about 60% - 70% in best configurations. The average F1 and average Recall values for best configurations of P2CE tested with Google Vision API and Microsoft Cognitive Services Vision API are shown in Fig. 10, 11, 12 and 13.

The best configurations identified for both Google Vision API and Microsoft Cognitive Services Vision API are Octagonal Prism Maps with Field of View of $45^{\circ} – 60^{\circ}$. The configuration with Field of View $52^{\circ}$ marginally outperforms $45^{\circ}$ and $60^{\circ}$ in overall. Use of Octagonal Prism Maps over Hexagonal Prism maps has an advantage of only about 3% when considering both parties best FOV angles. Thus, if computing power is a concern, a Hexagonal Prism Map with $60^{\circ}$ FOV should provide the best results. As shown in Fig. 10 and 11, the use of Cube Maps yield a lesser F1 score of about 60% when compared to 70% - 80% range of Octagonal Prism Maps and Hexagonal Prism Maps.

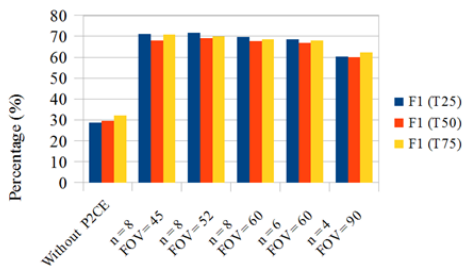

Fig. 10.   Average F1 with Google Vision API.

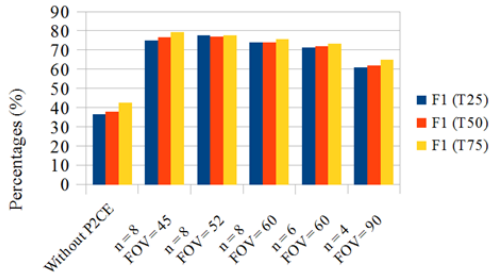

Fig. 11.   Average F1 with MS Cognitive Services API.

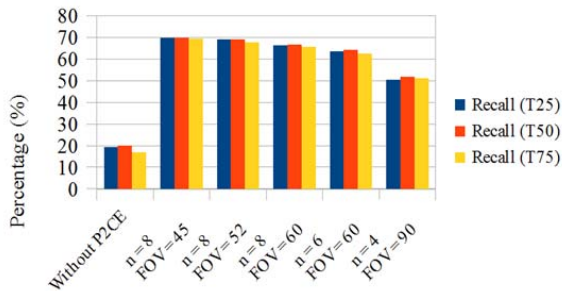

Fig. 12.   Average Recall with Google Vision API.

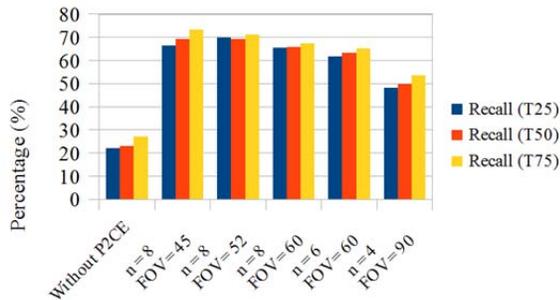

Fig. 13.   Average Recall with MS Cognitive Services API.

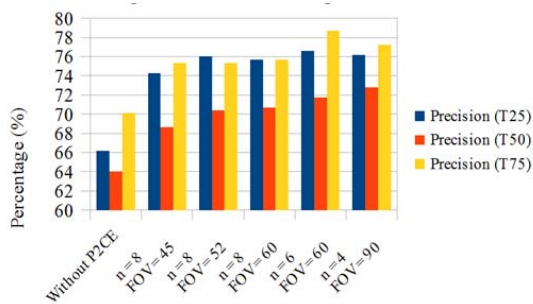

Fig. 14.   Average precision with Google Vision API.

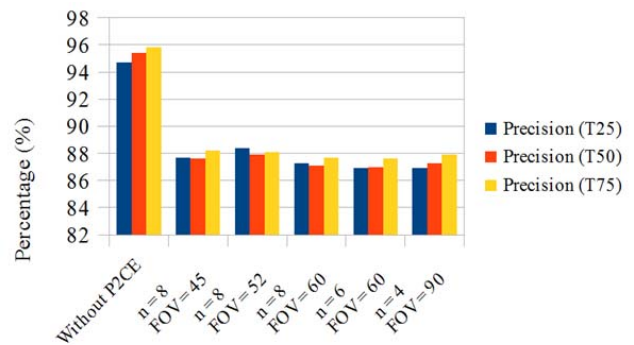

Fig. 15.   Average precision with MS Cognitive Services API.

The raw data of average figures represented in Fig. 10 to 15 are subjected to a Standard Deviation within range 10% - 20%. As evident in Fig. 15, the precision of figures from Ms. Cognitive Services Vision API drops to approximately 87.4% from 95.8% when P2CE is introduced, yet 87.4% may be considered as a healthy precision. This drop is mainly due to improvement in recall, which has made it possible for more labels to be identified. Additionally, in certain occasions, items with less significance are also identified with higher confidence due to effect of cutting photosphere into pieces. A portion of a photosphere is not always accurate in describing the entire picture. Similarly, as evident in Fig. 14, with the increase of $n$ and decrease of $FOV$, the resulting gradual increase in recall and the effects of cutting has caused precision to drop in Google Vision API as well.

Directly processing 2:1 photospheres through Cognitive Analysis APIs would only recognize very small objects closer to the center of the image or very large objects that contain randomized repetitive patterns. The very small objects tend to be less distorted by high Field of View Angle. Additionally, the center of 2:1 Image witnesses the least distortions. The very large objects with random repetitive patterns such as Sky and Water are also identifiable as their distortions would not destroy the identifiable visual features in them. However, the more important finer details in the picture such as Vehicles, Persons, Monuments or Buildings are not identifiable by directly processing the 2:1 Image. Nevertheless, by obtaining Prism Maps and processing their lateral faces as we proposed in this paper, the aforementioned can also be identified using typical Cognitive Image Analysis APIs. Consider the 2:1 Photosphere given in Fig. 9. Directly feeding the above image to Microsoft Cognitive Services Vision API yields labels *indoor*, *wall*, *room*, *living*, *ceiling*, *hotel*, *furniture*, *decorated,* and *flat* at $T_{25}$. However, by using an Octagonal Prism Map with FOV 52°, we are able to obtain additional labels such as *sofa*, *bedroom*, *curtain*, *door*, *floor*, *television*, *lamp*, *light* and *bed* at $T_{25}$. In certain rare occasions, directly processing 2:1 photospheres through typical Cognitive Image Analysis APIs can yield false positives. The curvy nature introduced to long objects by high field of view causes them to be falsely classified as objects that could curve such as trains or railway tracks.

## VI. Conclusion and Future Work

Giving cognition to photospheres is an area of research which has yet to be explored extensively. In this research we have introduced a methodology called P2CE to process 360 surround photos or photospheres using existing image cognition engines designed for conventional photos. We have introduced the novel concept of n-gonal Prism Maps, a generalization from Cube Maps as a method of improving accuracy in cognitive analysis of photospheres. Going further deep into the analysis we have determined the optimal values regarding *FOV* and *n* which are parameters for generating the Prism Maps. We have done rigorous testing on our findings and have improved accuracy (F1) up-to two times of the direct approach without P2CE.

The research we have conducted opens up many potentially viable research directions. A short-coming of current approach of P2CE is that it consumes about six times more computation than processing a conventional photo for cognition. Currently, we are working on reducing this short-come. An immediate future extension to the study can be on creating consistency between different confidence values/scores obtained for processing different lateral sides of the same Prism Map. Such consistency would be useful if a person wants to determine the most suitable label for a photosphere. However, the current work made by us is sufficient to generate keywords that would describe a Photosphere and such keywords can be used for applications such as Search Indexing of Photospheres. This methodology has been used in the commercial app called TravelSphere [26], which is already in the Google Play Store. The approach taken by us can be used in applications in addition to cognition. Standard problems such as Face Detection or Object Detection in photospheres can also be benefitted by using the same approach we have given; however, we encourage further research on those extending the concept presented in this work.